\documentclass[10pt,twocolumn,letterpaper]{article}

\usepackage{cvpr} 
\usepackage{times}
\usepackage{epsfig}
\usepackage{graphicx}
\usepackage{amsmath}
\usepackage{amssymb}
\usepackage{hyperref}

\usepackage{booktabs}
\usepackage{url}
\usepackage{algorithm}
\usepackage{algorithmic}
\usepackage{color,soul}
\usepackage{multirow}
\usepackage{xcolor}
\usepackage{colortbl}
\usepackage{pifont}
\usepackage{marvosym}    

\usepackage{fancyhdr}
\fancypagestyle{plain}{
    \fancyhf{} 
    \fancyfoot[L]{This paper is accepted in IJCB 2025} 
    \fancyfoot[R]{\thepage} 
}
\begin{document}

\title{Unified Knowledge Distillation Framework: Fine-Grained Alignment and Geometric Relationship Preservation for Deep Face Recognition}


\author{%
  Durgesh Mishra\textsuperscript{$\dagger$,~\Letter}, \quad 
  Rishabh Uikey\\ 
  Indian Institute of Science Education and Research, Bhopal, India\\
  {\tt\small \{durgesh.mishraa10, rishabhuikey\}@gmail.com}
}


\maketitle
\thispagestyle{empty}


\begingroup
  \renewcommand\thefootnote{\relax}
  \footnotetext{%
    \textsuperscript{$\dagger$}Part of this work was done as a Research Fellow at DRDO Young Scientist Laboratory – AI.  \quad
    \textsuperscript{\Letter} Corresponding author.}%
\endgroup

\begin{abstract}
Knowledge Distillation (KD) is crucial for optimizing face recognition models for deployment in computationally limited settings, such as edge devices. Traditional KD methods, such as Raw L2 Feature Distillation or Feature Consistency (FC) loss, often fail to capture both fine-grained instance-level details and complex relational structures, leading to suboptimal performance. We propose a unified approach that integrates two novel loss functions: Instance-Level Embedding Distillation (ILED) and Relation-Based Pairwise Similarity Distillation (RPSD). ILED focuses on aligning individual feature embeddings by leveraging a dynamic hard mining strategy, thereby enhancing learning from challenging examples. RPSD captures relational information through pairwise similarity relationships, employing a memory bank mechanism and a sample mining strategy. This unified framework ensures both effective instance-level alignment and preservation of geometric relationships between samples, leading to a more comprehensive distillation process. Our unified framework outperforms state-of-the-art distillation methods across multiple benchmark face recognition datasets, as demonstrated by extensive experimental evaluations. Interestingly, when using strong teacher networks compared to the student, our unified KD enables the student to even surpass the teacher’s accuracy.
\end{abstract}

\section{Introduction}

Face Recognition (FR) is integral to various security and authentication systems due to its quick and trustworthy identification performance. Earlier FR methods relied on traditional handcrafted features \cite{eigenfaces, belhumeur1997eigenfaces, liu2002gabor}, which typically lacked generalization. The advent of deep convolutional neural networks (CNNs) \cite{taigman2014deepface, parkhi2015deep} significantly boosted FR performance, replacing traditional feature extraction approaches and achieving SoTA results on benchmark datasets \cite{IJBB, IJBC}.

Furthermore, there is a growing trend towards deploying FR systems on low-computation edge devices \cite{caldeira2024modelcompressiontechniquesbiometrics}, such as smartphones, wearables, and IoT devices, driven by the increasing demand for privacy, real-time processing, and cost-efficiency.
However, this introduces significant challenges, as large architectures such as ResNet100 \cite{he2015deepresiduallearningimage}, Inception-ResNet-v2 \cite{inceptionv4inceptionresnetimpactresidual}, and SENet-154 \cite{SENets}, which achieve high accuracy, are incompatible with the limited computational resources and the restricted memory capabilities inherent in these devices. 

To address these challenges, model compression techniques, such as knowledge distillation (KD) \cite{hinton2015distillingknowledgeneuralnetwork}, pruning \cite{pruning1}, and quantization \cite{gong2014compressingdeepconvolutionalnetworks}, play a crucial role in optimizing FR systems for deployment on low-computation devices. This paper focuses specifically on the KD approach, in which a complex teacher model transfers its learned knowledge to a comparatively simpler student model. This knowledge transfer between the student and teacher models can occur in three main ways \cite{Gou_2021}: response-based, feature-based, and relation-based KD.

Traditional KD methods \cite{hinton2015distillingknowledgeneuralnetwork, Zhao} often rely on KL Divergence with Soft Logits as the primary distillation loss.
Recent studies have shown that using L2 loss on raw unnormalized feature embeddings, also called Raw L2 Feature Distillation \cite{Rethinking, romero2015fitnetshintsdeepnets, Luo_Zhu_Liu_Wang_Tang_2016}, and L2 loss on normalized feature embeddings, known as Feature Consistency (FC) loss \cite{ShrinkTeaNetML, couple_face, ECKD}, can significantly outperform KL‑divergence‑based methods for aligning feature embeddings.
The FC loss is a simple, straightforward way to align the student model representation with that of the teacher, but this direct approach may not effectively capture fine-grained information, leading to suboptimal performance. Moreover, recent research in metric learning \cite{harwood2017smartminingdeepmetric, 8953619, wu2018samplingmattersdeepembedding} emphasizes the importance of hard negative samples for enhancing the discriminative power of feature embeddings. Thus, a more refined approach that dynamically targets hard samples is needed for more effective knowledge transfer. 

Additionally, while FC loss facilitates feature alignment at the instance level, it fails to capture the relational information between different samples or the structural relationships learned by the teacher model. Existing methods \cite{RelationKD, darkrank} have attempted to address this limitation by incorporating relational knowledge into the distillation process, but these methods lack dynamic hard mining strategies \cite{Ge_Zhang_Liu_Hua_Zhao_Jin_Wen_2020, darkrank} or suffer from static mining processes with significant computational overhead \cite{couple_face}. Given these limitations, there is a need for more advanced KD methods that can effectively capture both fine-grained instance-level details and complex relational structures, ensuring robust performance even on resource-constrained edge devices. Our proposed approach introduces two novel loss functions: Instance-Level Embedding Distillation (ILED) and Relation-Based Pairwise Similarity Distillation (RPSD); designed to overcome the shortcomings of existing techniques. The primary contributions of the paper are as follows:
\begin{itemize}
    \item We propose a novel ILED loss function based on the rescaled softplus function, incorporating a dynamic sample mining strategy to effectively handle both easy and hard samples, ensuring better alignment between the student and teacher models.
    \item We introduce the RPSD loss function, which leverages pairwise similarity relationships to capture complex geometric relationships within the embedding space, enhancing the knowledge transfer by focusing on relational information between samples.
    \item Our approach integrates both ILED and RPSD losses into a unified framework, balancing fine-grained instance-level details with broader relational structures, leading to improved performance in knowledge distillation tasks.
\end{itemize}

The remainder of the paper is structured as follows: Section \ref{litreview} provides an overview of related work on FR losses and distillation methods for FR. Section \ref{methodology} describes the details of the proposed method. Section \ref{Expts} outlines the experimental setup and presents the results, comparing the proposed approach with existing methods. Finally, Section \ref{conclusion} concludes the paper.

\begin{figure*}[t]                        
  \begin{center}
    \includegraphics[width=0.90\textwidth]{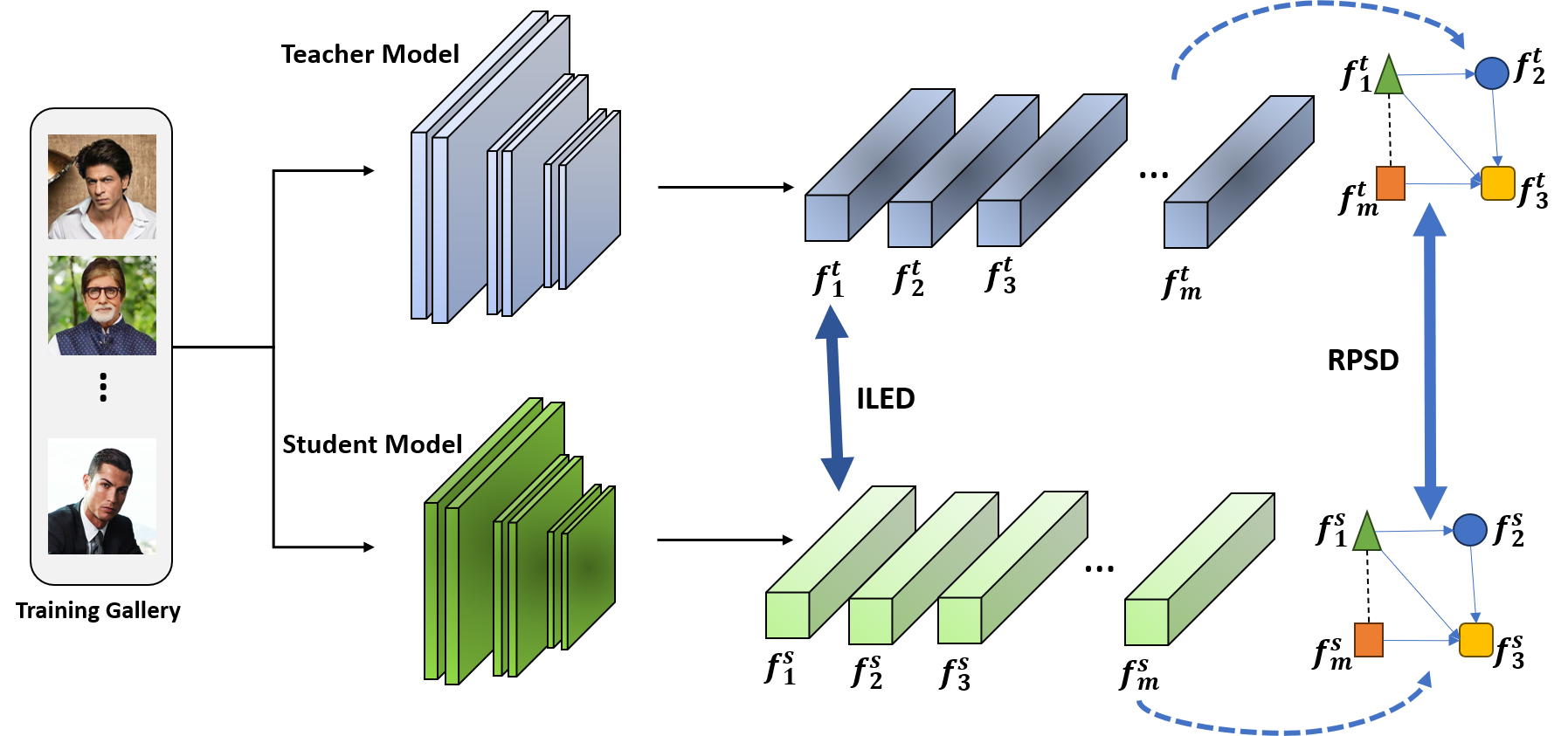}
  \end{center}
\caption{System overview of the unified knowledge distillation framework: Instance‑Level Embedding Distillation (ILED) dynamically aligns hard samples, while Relation‑Based Pairwise Similarity Distillation (RPSD) preserves geometric relationships through hard‑mining and a memory bank, jointly optimizing face recognition models.}

  \label{fig:sys_overview}
\end{figure*}



\section{Related Work}\label{litreview}

This section reviews FR literature and the KD methods applied to it.

\subsection{Face Recognition Losses} 

In FR, two widely used categories of loss functions are metric learning and angular softmax-based losses \cite{wang2022surveyfacerecognition}.
Some common losses used in metric learning are Contrastive Loss \cite{hadsell2006dimensionality} and Triplet Loss \cite{Schroff_2015, hermans2017defensetripletlossperson}. 
In addition to metric learning losses, FR models are also trained using softmax-based loss functions, which are treated as a classification problem. 
However, previous studies \cite{taigman2014deepface, 6909640} have shown that features learned through softmax loss are merely separable, lacking sufficient discriminative power. 
To overcome this limitation, L-Softmax \cite{liu2017largemarginsoftmaxlossconvolutional} was introduced aiming to push the decision boundaries between classes further apart in the loss function.  

Further, SphereFace \cite{liu2018spherefacedeephypersphereembedding} refines this by normalizing the weights and incorporating a more rigorous angular margin, creating a spherical decision boundary. CosFace \cite{wang2018cosfacelargemargincosine} simplifies this approach by using a cosine margin, directly optimizing the cosine similarity between features and class centers, which leads to better intra-class compactness. ArcFace \cite{deng2019arcface} extends these ideas further by applying an additive angular margin in the arccosine space, ensuring that features lie on a normalized hypersphere, which enhances both inter-class separability and intra-class compactness.

MagFace \cite{magface} and AdaFace \cite{adaface} both use feature norms as a proxy for image quality: higher norms mean high quality samples and lower norms mean low quality or hard samples. MagFace extends ArcFace with a feature norm-aware adaptive margin and a regularization term, encouraging high quality samples to have larger norms (closer to class centers) and pushing low quality ones away (smaller norms). AdaFace, by contrast, treats norms as fixed quality indicators for dynamic loss weighting, giving more weight to moderately hard but recognizable faces and less weight to unrecognizable ones without explicitly enforcing norm changes during backpropagation. SphereFace2 \cite{wen2022sphereface2binaryclassificationneed} emphasizes four key principles for binary classification-based training, transforming a k-class classification problem into k-binary classification tasks. A key component in the aforementioned methods is the normalization of features and weight vectors, which ensures that the learned features are distributed on a hypersphere, contributing to the robustness and effectiveness of the model.

\subsection{Knowledge Distillation in Face Recognition}
We now discuss the KD methods used for training the student model. 
DarkRank~\cite{darkrank} trains the student to mimic the teacher’s relative geometry within each mini-batch. For a query (the first sample in a mini-batch of size \(m\)), it measures Euclidean distances to the other \(m-1\) samples, converts them into scores, and sorts them to form the teacher’s ranking. The student does the same, and the loss forces its ranking to match the teacher’s: the soft version aligns the full distribution over all orderings, while the hard version only aligns the top-ranked list.

Grouped Knowledge Distillation~\cite{Zhao} splits teacher logits into Primary (most informative) and Secondary (less informative) groups, then uses a Binary‑KD loss on the Primary Group to align student distributions. Attention Similarity Knowledge Distillation (A-SKD) \cite{Shin} transfers attention maps from a high-resolution teacher network to a low-resolution student network using a Convolutional Block Attention Module (CBAM) \cite{cbam}.
Relational Knowledge Distillation (RKD) focuses on transferring the relational information between pairs or groups of data examples rather than individual outputs \cite{RelationKD}. This approach introduces two methods: Distance-wise Distillation and Angle-wise Distillation.

Correlation Congruence for Knowledge Distillation (CCKD) \cite{CCKD} improves student network performance by transferring instance-level information and correlations from a teacher network.
Exclusivity-Consistency Regularized Knowledge Distillation (EC-KD) \cite{ECKD} incorporates Weight Exclusivity Regularization (diversifying convolutional filters) and Feature Consistency Regularization (aligning intermediate features instead of logits) for FR.
ShrinkTeaNet~\cite{ShrinkTeaNetML} introduces Angular Distillation loss, where the student matches only the direction (not the magnitude) of each feature embedding with that of the teacher on a hypersphere. The KD loss is applied hierarchically at the final and intermediate layers with exponentially decaying weights.

CoupleFace~\cite{couple_face} combines Feature Consistency Distillation (FCD) and Mutual Relation Distillation (MRD) into a distillation loss. FCD uses the L2 distance on normalized feature embeddings to measure the difference between the teacher and student models. MRD captures mutual relation knowledge by evaluating the relative distances (cosine similarities) between pairs of samples. Hybrid Order Relational Knowledge Distillation (HORKD) \cite{Ge_Zhang_Liu_Hua_Zhao_Jin_Wen_2020} introduces a teacher stream trained on high-resolution images, a student stream learning to mimic the teacher on low-resolution images, and an assistant stream facilitating knowledge transfer. The method uses a loss function incorporating multiple levels of relational knowledge (1-order, 2-order, 3-order, and center-based). The approach in \cite{couple_face} assumes that FCD loss will rapidly align the student and teacher embeddings, but this convergence may be imperfect in complex datasets or limited training conditions. If convergence is inadequate, the mutual relation \( R(f^s_i, f^t_j) \) might not correctly approximate \( R(f^s_i, f^s_j) \), resulting in suboptimal knowledge transfer and performance. While the additional ``assistant'' stream in \cite{Ge_Zhang_Liu_Hua_Zhao_Jin_Wen_2020} improves knowledge transfer, it also introduces more complexity into the training process.

AdaDistill~\cite{AdaDistill} guides learning by matching each student embedding to an adaptive positive prototype: initially, the prototype closely resembles the teacher’s feature, encouraging sample-level imitation, but an exponential moving average gradually shifts it toward the teacher’s class centroid. Later training uses only the averaged prototype, ignoring intra-class variations (pose, age) and sample-to-sample relationships. 
Since the student trains solely with this centre-based margin loss~\cite{deng2019arcface, wang2018cosfacelargemargincosine}, without auxiliary cross-entropy, an inaccurate or shifting class centre may negatively impact learning, particularly for multi-modal classes or noisy data.

\section{Proposed Method}\label{methodology}

This section provides an overview of our unified KD framework in FR, as illustrated in Fig. \ref{fig:sys_overview}. Our approach combines Instance-Level Embedding Distillation and Relation-Based Pairwise Similarity Distillation to enhance the transfer of knowledge from a teacher model to a student model. Together, these methods aim to ensure more comprehensive and robust knowledge distillation. 

\subsection{Instance-Level Embedding Distillation (ILED)}

ILED focuses on individual examples within the training dataset. This involves training the student model to align its feature embeddings with those generated by the teacher model for each face image. For notations, let \( \mathbf{f}^t \in \mathbb{R}^d \) and \( \mathbf{f}^s \in \mathbb{R}^d \) denote the feature embeddings of the teacher and student models, where d represents the dimensionality of the embeddings. Traditional FR methods employ a FC loss \cite{ShrinkTeaNetML, couple_face}, shown in (\ref{eq:l2_norm}). The FC loss minimizes the L2 norm of the difference between the normalized teacher and student embeddings, enabling the student model to closely align its features with those of the teacher model.

Consider the L2 norm of the difference between the normalized teacher and student embeddings for a batch of size \( m \).

\begin{equation}
\mathcal{L}_{\text{FC}} = \frac{1}{m} \sum_{i=1}^{m} \left\| \frac{\mathbf{f}_i^t}{\|\mathbf{f}_i^t\|} - \frac{\mathbf{f}_i^s}{\|\mathbf{f}_i^s\|} \right\|^2
\label{eq:l2_norm}
\end{equation}

Upon simplification, this expression can be reformulated as
\begin{equation}
\mathcal{L}_{\text{FC}} = \frac{2}{m} \sum_{i=1}^{m} (1 - x_i).
\label{eq:loss_final}
\end{equation}

Where \( x_i \), representing the cosine similarity between the teacher and student embeddings for image \( i \), is given by
\begin{equation}
x_i = \frac{\mathbf{f}_i^t \cdot \mathbf{f}_i^s}{\|\mathbf{f}_i^t\| \, \|\mathbf{f}_i^s\|}.
\label{eq:cossine_sim}
\end{equation}

Equation (\ref{eq:loss_final}) provides a straightforward measure for aligning the student embeddings with the teacher embeddings.
However, such a direct approach may fail to capture the fine-grained information effectively, particularly when training a comparatively simpler student model. The simplistic use of cosine similarity as the loss function can overlook subtle nuances and intricate patterns present in the data, leading to suboptimal knowledge transfer. Several recent studies in metric learning \cite{harwood2017smartminingdeepmetric, 8953619, wu2018samplingmattersdeepembedding, adaface, magface} have shown that hard negative samples are important for improving the discriminative power of feature embeddings.
Therefore, a more refined approach is needed, one that incorporates additional strategies to dynamically focus on the most challenging examples, thereby ensuring a more comprehensive and accurate distillation of knowledge.

\vspace{2mm}
\textbf{Dynamic Sample Mining Strategy in ILED.}
To implement this refined approach, we employ a dynamic sample mining strategy that adjusts the contribution of each sample to the loss function based on its difficulty. Cosine similarity ranges from -1 to 1, with 1 indicating perfect alignment between the embeddings. Samples with high cosine similarity (close to 1) are considered \textit{``easy''} because they indicate that the student model's embeddings are already well-aligned with the teacher's embeddings. These easy samples result in a lower contribution to the loss, allowing the model to focus more on challenging examples. Conversely, samples with lower cosine similarity (further from 1) are considered \textit{``hard''} as they represent instances where the student model struggles to accurately replicate the teacher's embeddings. These hard samples contribute more significantly to the loss, directing the model's attention toward areas that require improvement.

To achieve this objective, we propose an ILED loss function that builds upon the rescaled softplus function \cite{pmlr-v15-glorot11a}, incorporating a dynamic easy/hard sample mining strategy, defined as:

{\small
\begin{equation}
\mathcal{L}_{\text{ILED}}
= \frac{1}{r}\ln\bigl(1 + \exp(-r(\bar{x} - s))\bigr)
  \sqrt{(\bar{x} - s)^2 + b}\,.
\label{eq:ikd_loss}
\end{equation}
}

Where $\bar{x}$ represents the average cosine similarity ($\cos$ $\theta$) between the student and teacher feature embeddings in a batch, given as

{\small
\begin{equation}
\bar{x} = \frac{1}{m}\sum_{i=1}^{m} x_i\,.
\label{eq:xmean}
\end{equation}
}

The hyperparameter \( s \) serves as a soft margin, providing a flexible target similarity that applies continuous penalties for deviations. Unlike a hard margin with a strict cutoff,  \( s \) smoothly guides the model toward the desired similarity. The hyperparameter r controls the steepness of the loss curve around the soft margin term  \( s \). Higher values of \( r \) make the function more sensitive to deviations from  \( s \), emphasizing hard samples. A small constant \( b \) is used to ensure numerical stability and maintain smoothness. To control the relative contributions of ILED loss, a hyperparameter \( \lambda_{\text{ILED}} \) is used. More importantly, ILED matches student and teacher embeddings solely through cosine similarity, so both vectors are L2‑normalised onto the unit hypersphere; alignment depends only on direction and ignores norms, granting the student extra degrees of freedom to interpret its teacher’s knowledge during learning.

\begin{figure}[t]                         
  \begin{center}
    \includegraphics[width=0.99\linewidth]{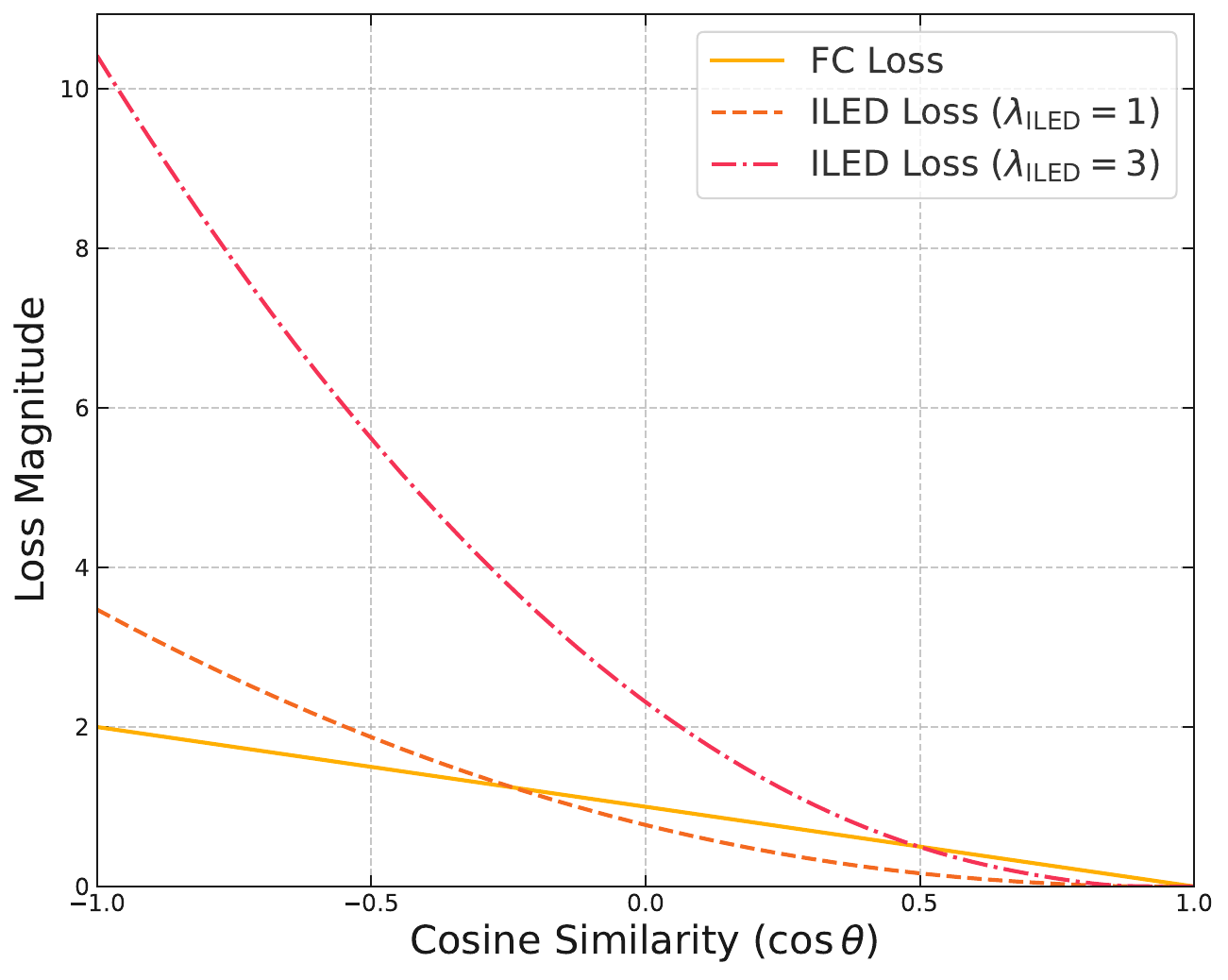}
  \end{center}
  \caption{The plot illustrates the FC loss and ILED losses at
    different scaling factors $\lambda_{\text{ILED}}$ while keeping
    the hyper-parameters fixed at $s = 0.85$, $b = 0.1$, and $r = 40$.}
  \label{fig:ILED_plot}
\end{figure}

Fig. \ref{fig:ILED_plot} shows a comparison between the traditional FC loss and the proposed ILED loss for varying scaling factors, \( \lambda_{\text{ILED}} \). The ILED loss uses hyperparameters such as \( r \) (steepness), \( s \) (soft margin), and \( b \) (smoothness) to dynamically adjust the loss based on sample difficulty. This approach emphasizes harder samples—those with lower cosine similarity, by assigning them a higher loss value, while easier samples—those with higher cosine similarity, contribute less to the overall loss. 
Overall, there are two main terms in (\ref{eq:ikd_loss}), a \textit{``logarithmic term''} and a \textit{``distance weighting component''}. The logarithmic term measures the misalignment between the average cosine similarity score $\bar{x}$ and a desired target value \(s\), smoothly penalizing deviations while allowing for gradient-based optimization. This term encourages the model to align its outputs closely with the target similarity. The distance weighting component dynamically adjusts the importance of each sample based on its distance from the target \(s\), amplifying the contribution of hard samples (those further from the target) and minimizing the impact of easy samples (those closer to the target). Together, these terms ensure a balanced focus, guiding the model to learn effectively by correcting difficult cases while not over-penalizing already well-aligned examples.

\subsection{Relation-Based Pairwise Similarity Distillation (RPSD)}

While ILED focuses on individual embeddings, RPSD takes a different approach by leveraging the relational structure between feature embeddings from a teacher model to a student model using the pairwise relationships between samples. This approach enhances knowledge transfer by matching the pairwise cosine similarities of teacher and student feature embeddings, ensuring the student learns the underlying geometric structure of the teacher’s embedding space.

\vspace{2mm}
\textbf{Pairwise Relations and Higher-Order Dependencies.} In angular margin-based loss functions, embeddings are normalized and projected onto a hypersphere while training \cite{wang2018cosfacelargemargincosine, deng2019arcface}. Since the feature embeddings are constrained to lie on the surface of a unit hypersphere, the angles (or cosine similarities) between them are sufficient to describe their geometric relationships. Thus, it is reasonable to assume that matching the cosine similarity relations between samples from the student and teacher models will effectively capture the underlying knowledge.

For normalized embeddings, higher-order relations (such as those between triplets) can be seen as combinations of pairwise relations, which simplifies the distillation process by focusing on matching these pairwise relations between the student and teacher models. To illustrate this, consider the example of a triplet relation involving three normalized vectors \( \mathbf{f}_t^i, \mathbf{f}_t^j, \mathbf{f}_t^k \in \mathbb{R}^d \), which represent feature embeddings from the teacher model. The angle between any two of these vectors can be expressed using pairwise cosine similarities:
\begin{equation}
\cos \theta_{ij} = \mathbf{f}_t^i \cdot \mathbf{f}_t^j, \quad \cos \theta_{jk} = \mathbf{f}_t^j \cdot \mathbf{f}_t^k, \quad \cos \theta_{ik} = \mathbf{f}_t^i \cdot \mathbf{f}_t^k,
\label{eq:pairwise_cosine}
\end{equation}

where \( \theta_{ij} \), \( \theta_{jk} \), and \( \theta_{ik} \) are the angles between the respective vectors.

To compute a higher-order relation, such as the angle formed by the three vectors, we consider the angle between the vectors \( \mathbf{f}_t^i - \mathbf{f}_t^j \) and \( \mathbf{f}_t^k - \mathbf{f}_t^j \). This can be expressed as:

\begin{equation}
\cos \angle \mathbf{f}_t^i \mathbf{f}_t^j \mathbf{f}_t^k = \frac{(\mathbf{f}_t^i - \mathbf{f}_t^j) \cdot (\mathbf{f}_t^k - \mathbf{f}_t^j)}{\| \mathbf{f}_t^i - \mathbf{f}_t^j \| \, \| \mathbf{f}_t^k - \mathbf{f}_t^j \|}.
\label{eq:cosine_angle}
\end{equation}

By expanding the dot product and utilizing the normalization of the vectors, we arrive at the following

\begin{equation}
(\mathbf{f}_t^i - \mathbf{f}_t^j) \cdot (\mathbf{f}_t^k - \mathbf{f}_t^j) = \cos \theta_{ik} - \cos \theta_{ij} - \cos \theta_{jk} + 1\,.
\label{eq:dot_product_expansion}
\end{equation}

The norms of the differences can be computed as
\begin{equation}
\begin{split}
\| \mathbf{f}_t^i - \mathbf{f}_t^j \| &= \sqrt{2(1 - \cos \theta_{ij})}, \\
\| \mathbf{f}_t^j - \mathbf{f}_t^k \| &= \sqrt{2(1 - \cos \theta_{jk})}.
\end{split}
\label{eq:norm_differences}
\end{equation}

Substituting expression \eqref{eq:dot_product_expansion} and \eqref{eq:norm_differences} into  \eqref{eq:cosine_angle} yields the final form:
\begin{equation}
\cos \angle \mathbf{f}_t^i \mathbf{f}_t^j \mathbf{f}_t^k = \frac{\cos \theta_{ik} - \cos \theta_{ij} - \cos \theta_{jk} + 1}{\sqrt{2(1 - \cos \theta_{ij})} \, \sqrt{2(1 - \cos \theta_{jk})}}.
\label{eq:final_cosine_angle}
\end{equation}

This example demonstrates that the angle formed by three normalized vectors, such as feature embeddings, can be entirely expressed using pairwise cosine similarities \( \cos \theta_{ij} \), \( \cos \theta_{jk} \), and \( \cos \theta_{ik} \). Therefore, matching the pairwise relationships between the student and teacher models should provide sufficient information for effective knowledge distillation. Moreover, since the vectors are normalized, calculating the pairwise similarity is simply done by computing the dot product between the matrices. This approach simplifies the calculations and is computationally efficient.

\vspace{2mm}
\textbf{Memory Bank Mechanism for RPSD.}
In transferring relational structure information from a teacher model to a student model, it is essential to capture the relationships between all possible pairs of samples in a dataset. However, computing all possible pairs in a dataset to capture relational information is computationally infeasible, especially in large datasets, due to the quadratic growth in the number of pairs as the dataset size increases. Consequently, the size of mini-batches imposes a practical limit on the extent of pairwise coverage that can be achieved. This constraint can lead to suboptimal learning of the student model, as the similarity information is only derived from a small set of examples at a time. To address this limitation, we employ a memory bank mechanism that maintains dynamic storage of feature embeddings using a queue and dequeue strategy \cite{he2020momentumcontrastunsupervisedvisual}. Specifically, we use a first-in-first-out (FIFO) queue where the oldest mini-batch is dequeued when the current mini-batch is enqueued. This mechanism allows the computation of similarities between the current mini-batch and a more extensive pool of data available in the memory bank.

Let the size of each mini-batch be \( m \), and the total capacity of the memory bank be \( q \). The embeddings of the teacher and student models stored in the memory banks are represented as \( \mathbf{F}_t = [\mathbf{f}_t^1, \mathbf{f}_t^2, \ldots, \mathbf{f}_t^q] \in \mathbb{R}^{q \times d} \) and \( \mathbf{F}_s = [\mathbf{f}_s^1, \mathbf{f}_s^2, \ldots, \mathbf{f}_s^q] \in \mathbb{R}^{q \times d} \), respectively, where \( d \) is the embedding dimension.

For each training iteration, we compute the cosine similarities between the embeddings of the current mini-batch and those stored in the memory bank for student and teacher models. Let the embeddings of the current mini-batch be denoted by \( \mathbf{E}_t = [\mathbf{e}_t^1, \mathbf{e}_t^2, \ldots, \mathbf{e}_t^m] \in \mathbb{R}^{m \times d} \) for the teacher model and \( \mathbf{E}_s = [\mathbf{e}_s^1, \mathbf{e}_s^2, \ldots, \mathbf{e}_s^m] \in \mathbb{R}^{m \times d} \) for the student model. The cosine similarity matrices are defined as:


\begin{equation}
\mathbf{S}_t(i, j) = \frac{\mathbf{e}_t^i \cdot (\mathbf{f}_t^j)^{\top}}{\|\mathbf{e}_t^i\| \|\mathbf{f}_t^j\|}, \quad \mathbf{S}_s(i, j) = \frac{\mathbf{e}_s^i \cdot (\mathbf{f}_s^j)^{\top}}{\|\mathbf{e}_s^i\| \|\mathbf{f}_s^j\|},
\label{eq:similarity}
\end{equation}

where \( \mathbf{S}_t \in \mathbb{R}^{m \times q} \) and \( \mathbf{S}_s \in \mathbb{R}^{m \times q} \) are the cosine similarity matrices of the teacher and student models, respectively. These matrices capture the pairwise cosine similarities between the embeddings in the current mini-batch and all the embeddings stored in the memory bank.

To quantify the difference in the relational feature spaces of the teacher and student models, the absolute element-wise difference between the corresponding similarity matrices is calculated.

\begin{equation}
\mathbf{D}(i, j) = \left| \mathbf{S}_t(i, j) - \mathbf{S}_s(i, j) \right| 
\label{eq:abs_diff}
\end{equation}

The resulting matrix \(\mathbf{D} \in \mathbb{R}^{m \times q}\) quantifies the dissimilarity between the teacher and student models based on their pairwise cosine similarities, capturing the alignment of their feature representations. The elements of $\mathbf{D}$ range from 0 to 2 but tend to be close to 0, even when the vectors are dissimilar, due to the properties of the cosine similarity difference. We use absolute differences instead of squared differences, as squaring smaller values would result in even smaller values, thus reducing the sensitivity of the measure to subtle variations between the models' embeddings.

The total dissimilarity across all pairs of embeddings is obtained by summing all elements of the matrix \(\mathbf{D}\).

\begin{equation}
\text{Total Dissimilarity} = \sum_{i=1}^{m} \sum_{j=1}^{q} \mathbf{D}(i, j) 
\label{eq:total_dissimilarity}
\end{equation}

To make the loss scale invariant and comparable across different batch sizes and memory bank capacities, the total dissimilarity is normalized by the number of elements (\(m \times q\)) in the matrix, resulting in the Normalized Dissimilarity, denoted as \(\Delta_{\text{norm}}\). This normalization produces a single scalar value that represents the average discrepancy between the similarity matrices of the student and teacher models, facilitating consistent and fair comparisons.

\vspace{2mm}
\textbf{Dynamic Sample Mining Strategy in RPSD.}
To effectively handle both easy and hard samples, we adopt a dynamic sample mining strategy, same as (\ref{eq:ikd_loss}), that adjusts the contribution of each sample to the loss based on its normalized dissimilarity value. A sample with \(\Delta_{\text{norm}}\) very close to 0 is considered \textit{``easy,''} indicating that the student model has effectively captured the teacher's relational structure and thus contributes less to the overall loss, allowing the model to prioritize more hard samples. Conversely, samples with \(\Delta_{\text{norm}}\) values that deviate further from 0 are considered \textit{``hard''} samples, representing instances where the student model struggles to replicate the relational structure of the teacher. These hard samples indicate areas where the student model has not yet learned effectively, and therefore, they contribute more significantly to the loss. This strategy ensures that the student model focuses on learning from the most challenging examples, driving improvement where it is most needed.

To dynamically adjust the impact of each sample based on its normalized similarity difference, we propose the RPSD loss function, capturing the relational structure between samples:

\begingroup
\small
\begin{equation}
\mathcal{L}_{\text{RPSD}} = \frac{1}{r'} \log\left(1 + \exp\left(r' \cdot (\Delta_{\text{norm}} - t)\right)\right) \cdot \sqrt{(\Delta_{\text{norm}} - t)^2 + b'}
\label{eq:rkd_loss}
\end{equation}
\endgroup

Where \(\Delta_{\text{norm}}\) represents the Normalized Dissimilarity across all samples, and \(r'\), \(t\), and \(b'\) are hyperparameters that control the shape and scale of the loss function. Here, \( t \) serves as a threshold or transition parameter that defines when the loss begins to increase significantly, distinguishing between \textit{``easy''} and \textit{``hard''} samples. The logarithmic term helps control or dampen the contribution of \textit{``easy''} samples, ensuring that the loss does not overly penalize the model for minor deviations. In contrast, the square root term enhances the emphasis on \textit{``hard''} samples, directing the model's learning focus toward areas where the alignment between the teacher and student embeddings is weaker. Together, these terms balance the loss function's response to both easy and hard samples. A hyperparameter \( \lambda_{\text{RPSD}} \) controls the relative contributions of the RPSD loss. 

\begin{figure}[t]                              
  \begin{center}
    \includegraphics[width=0.99\linewidth]{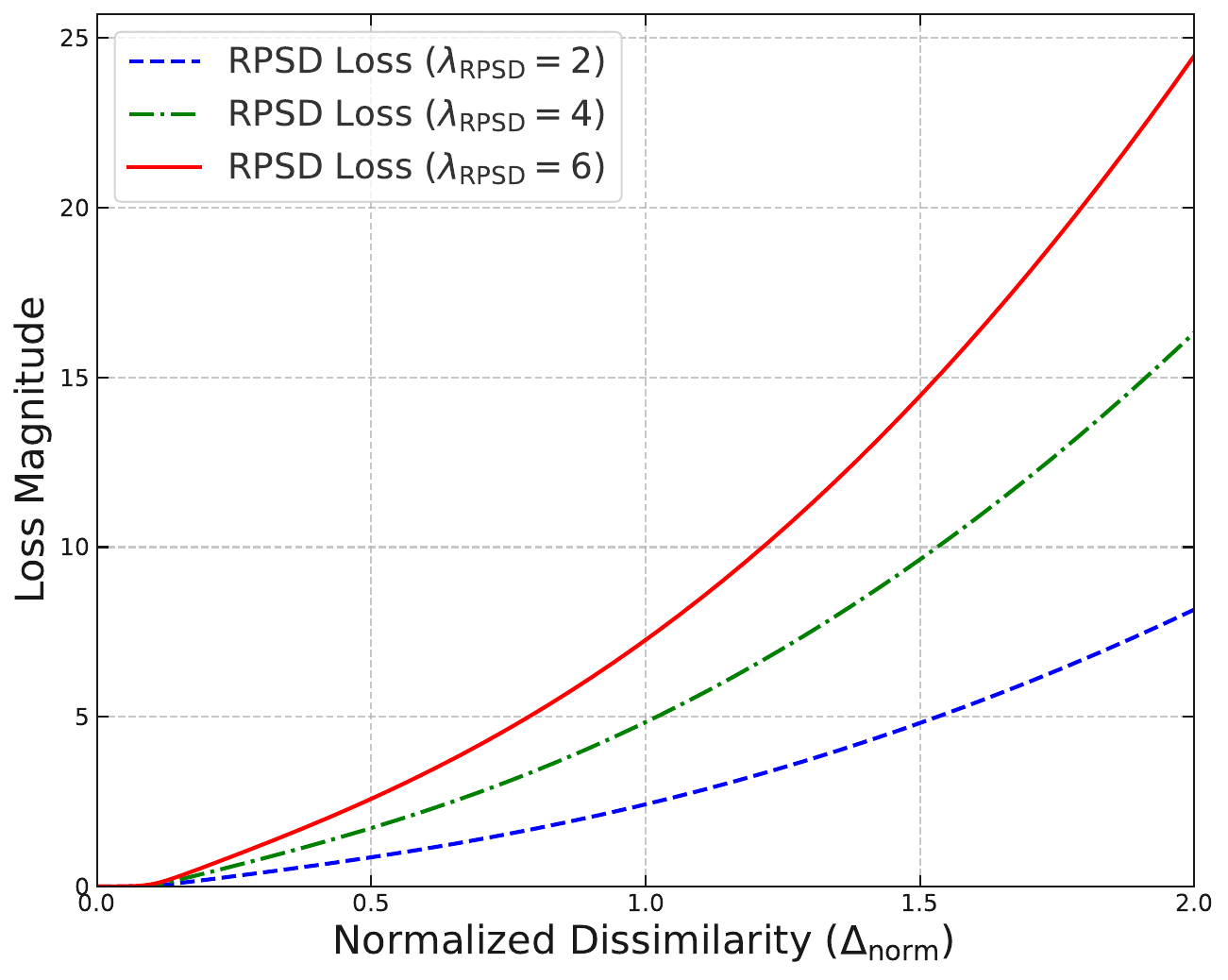}
  \end{center}
  \caption{The plot illustrates the RPSD loss at different scaling
    factors $\lambda_{\text{RPSD}}$ while the hyper-parameters are fixed
    to $t = 0.1$, $b' = 1$, and $r' = 60$.}
  \label{fig:RPSD_plot}
\end{figure}

Fig. \ref{fig:RPSD_plot} illustrates the behavior of the RPSD loss for different scaling factors, \( \lambda_{\text{RPSD}} \), across the range of normalized dissimilarity values from 0 to 2. The plot demonstrates how the RPSD loss function emphasizes hard samples (higher dissimilarity values) by increasing the loss magnitude while still allowing for dynamic adjustment based on the chosen scaling factor. This formulation ensures that the student model aligns its internal relational structure with that of the teacher model.

\subsection{Unified Loss Function}
The unified KD loss function for training the student model is defined as a weighted combination of the ILED loss, RPSD loss, and standard FR loss:

\begin{equation}
\mathcal{L}_{\text{total}} = \lambda_{\text{ILED}} \cdot \mathcal{L}_{\text{ILED}} + \lambda_{\text{RPSD}} \cdot \mathcal{L}_{\text{RPSD}} + \mathcal{L}_{\text{FR}}
\end{equation}

Where \( \mathcal{L}_{\text{ILED}} \) is the Instance-Level Knowledge Distillation loss as defined in \eqref{eq:ikd_loss}, \( \mathcal{L}_{\text{RPSD}} \) is the Relation-Based Knowledge Distillation loss as defined in \eqref{eq:rkd_loss}, and \( \mathcal{L}_{\text{FR}} \) is the standard FR loss, which helps the student model learn the correct class labels directly from the training data. The parameters \( \lambda_{\text{ILED}} \) and \( \lambda_{\text{RPSD}} \) are hyperparameters that control the relative contributions of each loss component.

Refer to Appendix A for a detailed, step‑by‑step pseudocode of our unified KD framework (ILED + RPSD). It details every stage of training, including KD loss calculations, memory‑bank updates, and parameter updates.


\section{Experiments and Results}\label{Expts}

\begin{table*}[t]
  \centering
    \caption{Consolidated performance comparison of knowledge-distillation methods.  
    Left: validation accuracy (\%) on LFW, AgeDB, CA-LFW, and CP-LFW (higher is better).  
    Right: verification rate (VR \%) on IJB-B and IJB-C at FAR = $10^{-5}$ and $10^{-4}$ (higher is better).  
    *Student model trained without any distillation, using only the SphereFace2 loss.}
  \label{tab:all_results}
  \begin{tabular}{lcccccccc}
    \toprule
    \multirow{2}{*}{Method} &
      \multicolumn{4}{c}{Validation accuracy (\%)} &
      \multicolumn{2}{c}{IJB‑B VR (\%)} &
      \multicolumn{2}{c}{IJB‑C VR (\%)} \\[2pt]
    \cmidrule(lr){2-5}\cmidrule(lr){6-7}\cmidrule(lr){8-9}
       & LFW & AgeDB & CA‑LFW & CP‑LFW
       & $10^{-5}$ & $10^{-4}$
       & $10^{-5}$ & $10^{-4}$ \\
    \midrule
    Student model* & 99.417 & 93.367 & 93.417 & 90.917 & 79.903 & 89.085 & 85.749 & 91.266 \\
    Teacher model  & 99.683 & 95.633 & 94.533 & 93.117 & 85.910 & 92.142 & 90.377 & 94.089 \\
    \midrule
    KL‑DIV \cite{hinton2015distillingknowledgeneuralnetwork}         & 99.450 & 92.517 & 93.317 & 89.800 & 77.945 & 87.215 & 83.438 & 89.692 \\
    Raw L2 \cite{Rethinking, romero2015fitnetshintsdeepnets}        & 99.383 & 93.167 & 93.533 & 89.783 & 77.556 & 87.546 & 83.876 & 89.876  \\
    FC loss \cite{couple_face, ECKD}       & 99.533 & 94.483 & 93.850 & 91.300 & 81.782 & 89.552 & 86.516 & 91.686 \\
    DarkRank \cite{darkrank}       & 99.550 & 94.167 & 93.683 & 90.883 & 79.659 & 88.695 & 85.514 & 91.190  \\
    ShrinkTeaNet \cite{ShrinkTeaNetML}  & 99.533 & 94.000 & 93.767 & 91.333 & 82.269 & 89.942 & 87.084 & 92.182  \\
    CCKD \cite{CCKD}           & 99.567 & 94.417 & 93.950 & 91.167 & 82.123 & 89.942 & 87.202 & 92.120 \\
    EC‑KD  \cite{ECKD}        & 99.583 & 93.283 & 93.583 & 90.800 & 79.474 & 89.114 & 85.478 & 91.251  \\
    ILED only      & 99.583 & 94.617 & \textbf{94.033} & 91.800 & 82.317 & 89.912 & 86.823 & 91.906 \\
    RPSD only      & 99.567 & 94.583 & 93.850 & 90.783 & 80.282 & 89.279 & 86.036 & 91.543 \\
    Unified KD    & \textbf{99.617} & \textbf{94.817} & 94.000 & \textbf{91.900}
                   & \textbf{82.405} & \textbf{90.117} & \textbf{87.288} & \textbf{92.458} \\
    \bottomrule
  \end{tabular}
\end{table*}

\textbf{Datasets.}
For training, a preprocessed VGGFace2 dataset \cite{VGGFace2} is used in all methods. In our experiments with MS1M‑V2, a refined version of the MS‑Celeb‑1M dataset \cite{guo2016msceleb1m, deng2019arcface} and VGGFace2, we observed that VGGFace2 converged faster and achieved a lower training loss in fewer iterations. This is likely due to its cleaner labels and greater intra-class variability, while MS1M-V2 requires more iterations due to its larger dataset size and higher noisy labels.
The preprocessed VGGFace2 dataset contains 3.1 million images from 8.6 thousand identities. The preprocessing process for the VGGFace2 dataset \cite{wen2022sphereface2binaryclassificationneed} involves cropping each face image using a similarity transformation based on five facial landmarks detected by MTCNN \cite{MTCNN}. This results in images of size 112 $\times$ 112 pixels. Additionally, each RGB pixel value, originally ranging from [0, 255], is normalized to the range of [$-1$, $1$].
For validation, four datasets are used: LFW \cite{LFW}, AgeDB \cite{AGEDB}, CA-LFW \cite{CALFW}, and CP-LFW \cite{CPLFWTech}, each containing 3,000 positive pairs and 3,000 negative pairs.
For testing, two large-scale IARPA Janus benchmarks are used: IJB‑B\cite{IJBB} (21,798 images and 55,026 frames from 7,011 videos, 1,845 identities) and its extension IJB‑C \cite{IJBC} (31,334 images and 117,542 frames from 11,779 videos, 3,531 identities).

\vspace{2mm}
\textbf{Experimental setting.}
All methods are implemented using the OpenSphere GitHub repository \cite{opensphere} and using the PyTorch framework \cite{paszke2019pytorch}. For a fair comparison, ResNet100 is used as the teacher's backbone, and ResNet18 is used as the student's backbone \cite{he2015deepresiduallearningimage}. SphereFace2 \cite{wen2022sphereface2binaryclassificationneed} is employed as the traditional FR loss to train both the student and the teacher models. In experimentation on the VGGFace2 dataset, SphereFace2 showed slightly better performance compared to other state-of-the-art loss functions, including CosFace \cite{wang2018cosfacelargemargincosine}, ArcFace\cite{deng2019arcface}, SphereFace\cite{liu2018spherefacedeephypersphereembedding}, and AdaFace\cite{adaface}.
The hyperparameters used for training are consistent with those reported in the paper: $\lambda$ = 0.7, r=40, m=0.4, t=3.0, where m is the CosFace-type additive margin. We use Stochastic Gradient Descent (SGD) \cite{polyak1964some} with a momentum of 0.9 and a batch size of 64. A Step Decay learning rate schedule is applied, starting at 0.1 and reducing by a factor of 10 at 50K, 100K, 120K, and 140K iterations. For testing, the model weights from the 140K iteration checkpoint are used. The experiments were conducted using NVIDIA RTX 4090 and RTX 6000 Ada GPUs to perform all computations.

\subsection{Results on the Validation and Test Datasets}

In this section, we validate the effectiveness of our proposed methods—ILED only, RPSD only, and Unified KD framework, by comparing them against several baseline methods commonly used in KD methods for FR tasks. We compare our method with KL Divergence with Soft Logits \cite{hinton2015distillingknowledgeneuralnetwork},  Raw L2 Feature Distillation (Raw L2) \cite{Rethinking, romero2015fitnetshintsdeepnets}, FC loss \cite{ShrinkTeaNetML, couple_face, ECKD}, DarkRank \cite{darkrank}, ShrinkTeaNet \cite{ShrinkTeaNetML}, Grouped Knowledge Distillation (GKD) \cite{Zhao}, Correlation Congruence KD (CCKD) \cite{CCKD}, Exclusivity-Consistency Regularized KD (EC-KD)\cite{ECKD}, and AdaDistill \cite{AdaDistill}. 
Refer to Appendix~C.3 for the full hyperparameter settings of all baseline KD methods.

For our proposed methods, the parameters for the ILED method were set as follows: \(r = 40\), \(s = 0.9\), \(b = 0.1\), and \( \lambda_{\text{ILED}} \) = 3. For the RPSD method, the parameters were configured with \(r' = 60\), \(t = 0.05\), \(b = 1\), and \( \lambda_{\text{RPSD}} \) = 40. The memory bank size \( q \) is set to be three times the batch size, and the feature embedding dimension for both the student and teacher models is 512.
The results on the validation datasets (LFW, AgeDB, CA-LFW, and CP-LFW) and on the test datasets (IJB-B and IJB-C) for the aforementioned methods are all shown in Table \ref{tab:all_results}. The numbers for the GKD and AdaDistill methods were very low, so we did not include them in the comparison table. 
AdaDistill is specifically designed for margin‑penalty Softmax losses (e.g., ArcFace and CosFace), making it incompatible with SphereFace2’s binary angular‑margin loss. Directly applying AdaDistill to SphereFace2 yields suboptimal results.

The results in Table \ref{tab:all_results} show that the unified approach of the ILED + RPSD methods consistently outperforms other knowledge-distillation techniques across multiple datasets. Specifically, it achieves the highest accuracy on LFW (99.617 \%), AgeDB (94.817 \%), and CP-LFW (91.900 \%), closely matching the teacher-model performance, and it delivers the best verification rates at lower false-accept rates on both IJB-B and IJB-C, demonstrating its robustness and effectiveness compared with the student model without distillation and with other methods. Overall, this unified approach proves to be the most effective. 

We conducted a small ablation study to evaluate the contributions of ILED and RPSD methods within our unified KD framework; see Appendix B. In Appendix C.1, we present results for different teacher–student pairings, illustrating how the capacity gap between models affects knowledge transfer. Appendix C.2 then examines the role of the teacher’s pretraining loss. Together, these analyses demonstrate the robustness of our method across both architectural and loss‑function variations.


\section{Conclusion}\label{conclusion}

In this paper, we propose a unified framework for KD in deep FR that combines two novel loss functions: ILED and RPSD. ILED dynamically focuses on challenging samples for better alignment, while RPSD captures relational structures to enhance the student's understanding of geometric relationships. Experiments on multiple benchmark datasets showed that our unified KD approach outperforms traditional FR methods, but the need to tune several hyperparameters is a limitation. Future work could focus on making the hyperparameters adaptive during training and the extension of our unified framework to other metric‑learning tasks such as Person Re‑Identification and Fine‑Grained Image Retrieval.


{\small
\bibliographystyle{ieee}
\bibliography{egbib}
}


\clearpage
\onecolumn
\appendix
\renewcommand{\thesection}{\Alph{section}}
\renewcommand{\thesubsection}{\Alph{section}.\arabic{subsection}}

\begin{center}
\section*{Appendix}
\end{center}

\section{Pseudo-Algorithm}\label{app:pseudo}

We present the pseudo-algorithm (Algorithm \ref{alg:training}) for our proposed Unified Knowledge Distillation (KD) framework for face recognition. The algorithm outlines the step-by-step process of our approach to optimize the student's model performance. In this pseudo-algorithm, we assume that the teacher model is already trained using a standard face recognition loss function (SphereFace2).
In the notation used, $r$, $s$, and $b$ are hyperparameters for the ILED, while $r'$, $t$, and $b'$ are for the RPSD. The parameters $\lambda$ and $\lambda'$ are weight factors that control the contributions of the ILED and RPSD losses to the total KD loss. The size of the memory banks is denoted by $q$, and the updating of a new batch in the memory bank is done using a queue and dequeue strategy, specifically employing a first-in-first-out (FIFO) method.

\begin{algorithm}[h]
\caption{Unified Knowledge Distillation Framework for Deep Face Recognition}
\label{alg:training}
\begin{algorithmic}[1]

\STATE \textbf{Input:} Student model $S$, Teacher model $T$, Training data $\mathcal{D} = \{x_i, y_i\}_{i=1}^N$, Hyperparameters $(\lambda, r, b, s, \lambda', r', b', t, q)$
\STATE \textbf{Output:} KD-optimized student model $S$

\STATE \textbf{Procedure:}
\STATE Initialize parameters for $S$ and $T$
\STATE Load pretrained weights for $T$ and freeze its parameters
\STATE Set student and teacher memory banks: $F_s \gets \text{None}, \; F_t \gets \text{None}$
\STATE Initialize $L_{\text{RPSD}} \gets 0$

\FOR{each iteration in the training process}
    \STATE Sample mini-batch $(\{x_i\}, \{y_i\}) \subset \mathcal{D}$, where $i = 1, \ldots, m$

    \STATE Compute student and teacher embeddings for mini-batch: $E_s \gets S(\{x_i\}), \; E_t \gets T(\{x_i\})$

    \STATE Compute recognition loss for student: $L_{\text{student}}$
    
    \STATE Calculate cosine similarities: $x_i \gets \cos(f_s^i, f_t^i), \; \forall (f_s^i, f_t^i) \in (E_s, E_t)$

    \STATE \textbf{Compute ILED Loss:}
    \[
    L_{\text{ILED}} = \frac{1}{m} \sum_{i=1}^{m} \frac{1}{r} \ln \left(1 + \exp(-r \cdot (x_i - s))\right) \cdot \sqrt{(x_i - s)^2 + b}
    \]

    \IF{memory banks are sufficiently populated}
        \STATE Compute teacher and student similarity matrices: $S_t \gets \cos(E_t, F_t^T), \; S_s \gets \cos(E_s, F_s^T)$
        \STATE Calculate pairwise similarity difference: $D \gets |S_t - S_s|$

        \STATE Normalize dissimilarity:
        \[
        \Delta_{\text{norm}} = \frac{1}{q \cdot m} \sum_{i=1}^{q} \sum_{j=1}^{m} D(i, j)
        \]
       
        \STATE \textbf{Compute RPSD Loss:}
        \[
        L_{\text{RPSD}} = \frac{1}{r'} \log \left(1 + \exp(r' \cdot (\Delta_{\text{norm}} - t)) \right) \cdot \sqrt{(\Delta_{\text{norm}} - t)^2 + b'}
        \]
    \ENDIF

    \STATE \textbf{Total Loss:}
    \STATE $L_{\text{KD}} \gets \lambda L_{\text{ILED}} + \lambda' L_{\text{RPSD}}$
    \STATE $L_{\text{total}} \gets L_{\text{student}} + L_{\text{KD}}$

    \STATE \textbf{Update Student Model:}
    \STATE Backpropagate and update parameters: $\theta_s \leftarrow \theta_s - \eta \nabla_{\theta_s} L_{\text{total}}$
    \STATE Update memory banks: $F_s \leftarrow \text{FIFO}(F_s, E_s), \; F_t \leftarrow \text{FIFO}(F_t, E_t)$

\ENDFOR

\RETURN optimized student model $S$

\end{algorithmic}
\end{algorithm}

\clearpage

\section{Ablation Study}

We conducted a small ablation study to assess the impact of each component in our Unified Knowledge Distillation framework by isolating Instance-Level Embedding Distillation (ILED) and Relation-Based Pairwise Similarity Distillation (RPSD). This study aims to discern the contribution of each method to the overall performance of the student model across multiple benchmark face recognition datasets. 

We assessed several variations of our approach on four validation datasets: LFW, AgeDB, CA-LFW, and CP-LFW. In accordance with the experimental setup outlined in the paper, a ResNet100 serves as the teacher backbone, while a ResNet18 is utilized as the student backbone. The SphereFace2 hyperparameters for training were kept consistent with those reported: $\lambda = 0.7$, $r = 40$, $m = 0.4$, $t = 3.0$, where $m$ represents the CosFace-type additive margin.

\subsection{Instance-Level Embedding Distillation (ILED)}

The proposed ILED method uses three main hyperparameters: $r$ (steepness), $s$ (soft target), and $\lambda_{\mathrm{ILED}}$ (weight).  The parameter $b$ (small positive constant) is kept constant at 0.1 throughout the experiments, and $\lambda_{\mathrm{ILED}}$ is set to 3 to maintain a balanced contribution to the overall loss. The performance of the ILED method with different configurations of these hyperparameters is presented in Table \ref{table:ILED_Performance}.

\begin{table}[h]
  \centering
  \caption{Performance of ILED with Different Hyperparameters on Validation Datasets}
  \label{table:ILED_Performance}
  \begin{tabular}{@{}ccc|cccc@{}}
    \toprule
    $\lambda_{\mathrm{ILED}}$ & $r$ & $s$ & LFW     & AgeDB   & CA-LFW  & CP-LFW  \\
    \midrule
      3 & 40 & 0.5 & 99.467 & 93.933 & 93.850 & 91.000 \\
      3 & 40 & 0.6 & 99.550 & 94.000 & 93.850 & 91.017 \\
      3 & 40 & 0.7 & 99.517 & 93.783 & 93.667 & 91.000 \\
      3 & 40 & 0.8 & \textbf{99.617} & 94.417 & 93.733 & 91.317 \\
      3 & 40 & 0.9 & 99.583 & \textbf{94.617} & \textbf{94.033} & \textbf{91.800} \\
    \midrule
      3 & 20 & 0.9 & 99.517 & 94.333 & 93.883 & 91.283 \\
      3 & 40 & 0.9 & \textbf{99.583} & \textbf{94.617} & \textbf{94.033} & \textbf{91.800} \\
      3 & 60 & 0.9 & 99.517 & 94.300 & 93.833 & 91.183 \\
    \bottomrule
  \end{tabular}
\end{table}

\noindent From Table \ref{table:ILED_Performance}, it is observed that the optimal performance for ILED is achieved with $s=0.9$ and $r=40$, suggesting that a higher soft target $s$ improves the alignment between the student and teacher embeddings.

\subsection{Relation-Based Pairwise Similarity Distillation (RPSD)}

The proposed RPSD function uses three main hyperparameters: $r'$ (steepness), $t$ (transition parameter), and $\lambda_{\mathrm{RPSD}}$ (weight).  The parameter $b$ (small positive constant) is kept constant at 1 throughout the experiments, and $\lambda_{\mathrm{RPSD}}$ is set to 60 to maintain a balanced contribution to the overall loss. The performance of the RPSD method with different configurations of these hyperparameters is presented in Table \ref{table:RPSD_performance}.

\begin{table}[h]
  \centering
  \caption{Performance of RPSD with Different Hyperparameters on Validation Datasets}
  \label{table:RPSD_performance}
  \begin{tabular}{@{}ccc|cccc@{}}
    \toprule
    $\lambda_{\mathrm{RPSD}}$ & $r'$ & $t$ & LFW     & AgeDB   & CA-LFW  & CP-LFW  \\
    \midrule
      60 & 60 & 0.05 & \textbf{99.517} & \textbf{94.167} & \textbf{93.717} & \textbf{90.867} \\
      60 & 60 & 0.10 & 99.433 & 93.567 & 93.433 & 90.600 \\
      60 & 60 & 0.15 & 99.417 & 93.100 & 93.400 & 90.750 \\
      60 & 60 & 0.20 & 99.450 & 93.133 & 93.417 & 90.717 \\
    \midrule
      60 & 40 & 0.05 & 99.533 & 94.150 & 93.633 & \textbf{91.183} \\
      60 & 60 & 0.05 & 99.517 & \textbf{94.167} & \textbf{93.800} & 90.867 \\
      60 & 80 & 0.05 & \textbf{99.550} & 93.733 & 93.717 & 91.000 \\
    \bottomrule
  \end{tabular}
\end{table}

\clearpage
\section{Experimentations}\label{sec:discussion}

In this section, we present a comprehensive evaluation of our Unified Knowledge Distillation (Unified KD) framework across a variety of settings. We first examine the impact of different teacher–student capacity gaps in Section C.1, then analyze the effect of the teacher’s pretraining loss in Section C.2, and provide implementation details for the baseline methods in Section C.3.

\subsection{Performance Across Teacher-Student Combinations}

Table~\ref{tab:teacher_student} compares students trained without any
distillation against our Unified KD method—which combines ILED and RPSD, across three teacher architectures (ResNet100, ResNet50, DPN98) and four
student backbones (ResNet18, MobileFaceNet, ResNet34, ResNet50).
All networks were trained using the SphereFace2 loss with the same hyperparameters as in the main paper (\(\lambda=0.7\), \(r=40\), \(m=0.4\) for the CosFace‐style additive margin, and \(t=3.0\)).  For our Unified KD experiments, we set the ILED parameters to \(r=40\), \(s=0.9\), \(\lambda_{\mathrm{ILED}}=9\), and the RPSD parameters to \(r'=60.0\), \(t=0.05\), \(\lambda_{\mathrm{RPSD}}=40.0\), using a memory‐bank size of \(q=3\).  All results were reported using model weights saved at 140k training iterations with a batch size of 64.  

\begin{table*}[ht]
  \centering
  \setlength{\tabcolsep}{5pt}
  \renewcommand{\arraystretch}{1.15}
   \caption{Teacher–student distillation matrix. The baseline without knowledge distillation is denoted by (\ding{55}), and Unified KD (ILED + RPSD) is denoted by (\ding{51}). All numbers represent accuracies (\%) on the LFW, AgeDB, CA‑LFW, and CP‑LFW datasets.}
  \label{tab:teacher_student}
  \begin{tabular}{llccccc}
    \toprule
    Teacher      & Student       & Scheme    & LFW     & AgeDB   & CA‑LFW  & CP‑LFW  \\
    \midrule

    \multirow{9}{*}{ResNet100}
      & --            & Teacher Only & 99.633  & 95.550  & 94.500  & 92.950  \\
    \arrayrulecolor{gray!50}\cmidrule(lr){2-7}\arrayrulecolor{black}
      & ResNet18      & \ding{55}  & 99.417  & 93.367  & 93.417  & 90.917  \\
      &               & \ding{51}  & 99.600  & 94.700  & 93.900  & 91.667  \\
    \arrayrulecolor{gray!50}\cmidrule(lr){2-7}\arrayrulecolor{black}
      & MobileFaceNet & \ding{55}  & 98.917  & 88.683  & 91.433  & 87.300  \\
      &               & \ding{51}  & 99.417  & 92.650  & 93.217  & 90.117  \\
    \arrayrulecolor{gray!50}\cmidrule(lr){2-7}\arrayrulecolor{black}
      & ResNet34      & \ding{55}  & 99.617  & 95.017  & 94.217  & 92.333  \\
      &               & \ding{51}  & 99.650  & 95.533  & 94.283  & 93.183  \\
    \arrayrulecolor{gray!50}\cmidrule(lr){2-7}\arrayrulecolor{black}
      & ResNet50      & \ding{55}  & 99.583  & 95.233  & 94.300  & 92.600  \\
      &               & \ding{51}  & 99.633  & 95.850  & 94.417  & 93.550  \\
    \midrule

    \multirow{9}{*}{ResNet50}
      & --            & Teacher Only & 99.583  & 95.233  & 94.300  & 92.733  \\
    \arrayrulecolor{gray!50}\cmidrule(lr){2-7}\arrayrulecolor{black}
      & ResNet18      & \ding{55}  & 99.417  & 93.367  & 93.417  & 90.917  \\
      &               & \ding{51}  & 99.550  & 94.600  & 93.933  & 91.767  \\
    \arrayrulecolor{gray!50}\cmidrule(lr){2-7}\arrayrulecolor{black}
      & MobileFaceNet & \ding{55}  & 98.917  & 88.683  & 91.433  & 87.300  \\
      &               & \ding{51}  & 99.417  & 92.717  & 93.100  & 90.100  \\
    \arrayrulecolor{gray!50}\cmidrule(lr){2-7}\arrayrulecolor{black}
      & ResNet34      & \ding{55}  & 99.617  & 95.017  & 94.217  & 92.333  \\
      &               & \ding{51}  & 99.667  & 95.417  & 94.267  & 92.883  \\
    \midrule

    \multirow{9}{*}{DPN98}
      & --            & Teacher Only & 99.233  & 90.617  & 92.550  & 90.083  \\
    \arrayrulecolor{gray!50}\cmidrule(lr){2-7}\arrayrulecolor{black}
      & ResNet18      & \ding{55}  & 99.417  & 93.367  & 93.417  & 90.917  \\
      &               & \ding{51}  & 99.500  & 92.500  & 93.267  & 90.800  \\
    \arrayrulecolor{gray!50}\cmidrule(lr){2-7}\arrayrulecolor{black}
      & MobileFaceNet & \ding{55}  & 98.917  & 88.683  & 91.433  & 87.300  \\
      &               & \ding{51}  & 99.383  & 90.750  & 92.417  & 90.083  \\
    \arrayrulecolor{gray!50}\cmidrule(lr){2-7}\arrayrulecolor{black}
      & ResNet34      & \ding{55}  & 99.617  & 95.017  & 94.217  & 92.333  \\
      &               & \ding{51}  & 99.667  & 93.717  & 93.700  & 92.550  \\
    \arrayrulecolor{gray!50}\cmidrule(lr){2-7}\arrayrulecolor{black}
      & ResNet50      & \ding{55}  & 99.583  & 95.233  & 94.300  & 92.600  \\
      &               & \ding{51}  & 99.650  & 93.917  & 93.817  & 92.800  \\
    \bottomrule
  \end{tabular}
\end{table*}

When the teacher is strong (e.g.\ ResNet100) compared to the student, Unified KD delivers
substantial improvements over without KD method, demonstrating that ILED + RPSD effectively transfers richer teacher embeddings into a smaller model. Interestingly, in some pairings, the Unified KD student even surpasses the teacher’s native performance, suggesting that the distillation losses also regularize and smooth teacher overconfidence.
Conversely, when the teacher is weaker than the student (for example, DPN98 teaching ResNet18), applying Unified KD sometimes yields worse results than training without distillation.
This indicates that when the teacher’s embeddings are less discriminative than the student’s capacity, the additional distillation losses can overconstrain the student and lead to suboptimal fitting.

\subsection{Effect of Teacher Loss Function on Student Performance}

In Table \ref{table:different_teachers}, the teacher model consistently uses the ResNet100 architecture with pretrained weights from an open-source repository, each trained with different FR losses. The student model uses ResNet18 trained with SphereFace2 as the FR loss. The proposed KD method maintains consistency across pretrained models and losses. Note that the lower accuracies for the SphereFace2‑pretrained teacher are because it was trained for significantly fewer iterations than the other open-source teacher models.

\begin{table}[ht]
  \centering
  \setlength{\tabcolsep}{4pt}    
  \caption{Performance of the student model trained with pretrained teacher models using various face recognition (FR) loss functions, evaluated on multiple validation datasets.}
  \label{table:different_teachers}
  \begin{tabular}{@{}lcccc@{}}
    \toprule
    \textbf{FR Loss}   & \textbf{LFW}  & \textbf{AgeDB} & \textbf{CA‑LFW} & \textbf{CP‑LFW} \\
    \midrule
    ArcFace            & 99.517        & 94.350         & 93.850          & 91.383          \\
    AdaFace            & 99.500        & 95.567         & 94.500          & 91.733          \\
    SphereFace2        & 99.417        & 93.367         & 93.417          & 90.917          \\
    \bottomrule
  \end{tabular}
  \normalsize                    
\end{table}

\subsection{Baseline Implementation Details}

For the L2 raw features and FC loss, we set the loss weights to $\lambda_{L2}=0.1$ (to compensate for the large L2 values) and $\lambda_{FC}=3$. In Standard Temperature‐Scaled KD, we used a temperature of $T=3$. For GKD, we chose $\tau=0.93$, $\lambda_1=8$, $\lambda_2=1$, and $T=1$. For ShrinkTeaNet Angular KD, we assigned the final‑layer weight $\lambda_{n} = 1$ and recursively defined each intermediate weight by $\lambda_{i} = \tfrac{1}{2}\,\lambda_{i+1}$ for $i = n-1,\dots,1$. DarkRank employs a scale factor $\alpha=3$, exponent $\beta=3$, and a list‐wise KL loss weight of $\lambda=0.1$.  For EC-KD, we used $\lambda_1$=0.0005 and  $\lambda_2$= 2 $\times$  $\lambda_1$, and due to the very small magnitude of the FC loss (in the range of $10^{-10}$), we normalized the features before computing the loss. Finally, in Correlation Congruence KD (CCKD), we used $\gamma=0.5$, $\beta=10$, $\alpha=0.5$, and replaced KL divergence with an L2‐mimic loss as recommended in the paper. Additionally, we fine-tuned the hyperparameters of these methods to further enhance their performance.

\end{document}